\documentclass[runningheads]{llncs}
\usepackage[T1]{fontenc}
\usepackage{graphicx}
\usepackage{textcomp}
\usepackage{color,xcolor}
\usepackage{caption}

\usepackage{subfigure}
\usepackage{booktabs}
\usepackage{hyperref}
\usepackage[switch]{lineno}
\usepackage{soul}
\usepackage{utfsym}
\usepackage{indentfirst}
\usepackage{times}
\usepackage{latexsym}

\usepackage{multirow}%
\usepackage{amsmath,amssymb,amsfonts}%
\usepackage{mathrsfs}%
\usepackage[title]{appendix}%
\usepackage{textcomp}%
\usepackage{manyfoot}%
\usepackage{booktabs}%
\usepackage{algorithm}%
\usepackage{algorithmicx}%
\usepackage{algpseudocode}%
\usepackage{listings}%
\begin{document}
\title{Improving monotonic optimization in heterogeneous multi-agent reinforcement learning with optimal marginal deterministic policy gradient}
%
%

\author{Xiaoyang Yu\inst{1,2}\orcidID{0000-0002-3592-5746} \and
Youfang Lin\inst{1,2} \and
Shuo Wang\inst{1,2} \and
Sheng Han\inst{1,2}}
\institute{School of Computer Science and Technology, Beijing Jiaotong University, Beijing 100044, China \and Beijing Key Laboratory of Traffic Data Mining and Embodied Intelligence,  Beijing 100044, China \\
\email{xiaoyang.yu@bjtu.edu.cn} \\
\email{shhan@bjtu.edu.cn}}

\maketitle              
\begin{abstract}
In heterogeneous multi-agent reinforcement learning (MARL), achieving monotonic improvement plays a pivotal role in enhancing performance.
The HAPPO algorithm proposes a feasible solution by introducing a sequential update scheme, which requires independent learning with No Parameter-sharing (NoPS). However, heterogeneous MARL generally requires Partial Parameter-sharing (ParPS) based on agent grouping to achieve high cooperative performance. Our experiments prove that directly combining ParPS with the sequential update scheme leads to the policy updating baseline drift problem, thereby failing to achieve improvement.
To solve the conflict between monotonic improvement and ParPS, we propose the Optimal Marginal Deterministic Policy Gradient (OMDPG) algorithm.
First, we replace the sequentially computed $Q_{\psi}^s(s,a_{1:i})$ with the Optimal Marginal Q (OMQ) function $\phi_{\psi}^*(s,a_{1:i})$ derived from Q-functions. This maintains MAAD's monotonic improvement while eliminating the conflict through optimal joint action sequences instead of sequential policy ratio calculations.
Second, we introduce the Generalized Q Critic (GQC) as the critic function, employing pessimistic uncertainty-constrained loss to optimize different Q-value estimations. This provides the required Q-values for OMQ computation and stable baselines for actor updates.
Finally, we implement a Centralized Critic Grouped Actor (CCGA) architecture that simultaneously achieves ParPS in local policy networks and accurate global Q-function computation. Experimental results in SMAC and MAMuJoCo environments demonstrate that OMDPG outperforms various state-of-the-art MARL baselines.

\end{abstract}
\section{Introduction}
In recent years, Multi-Agent Reinforcement Learning (MARL) has demonstrated significant potential \cite{zhu2024survey} in complex scenarios, including intelligent transportation systems \cite{APPI-qiao2023traffic,traffic+wu2020multi},
power control in wireless networks \cite{amorosa2023multi}, multi-agent air combat \cite{NCA-sun2023multi},
and smart grids \cite{FedMRL-sahoo2024fedmrl}.
It has been proved that the monotonic improvement guarantee is critical for the convergence of MARL policies \cite{mapg+bono2018cooperative,mappo+yu2021surprising}.
As a remarkable state-of-the-art method, HAPPO \cite{happo+kuba2021trust} implements monotonic improvement via the Multi-Agent Advantage Decomposition (MAAD) theorem with the sequential update scheme and independent policies.


However, implementing HAPPO into heterogeneous MARL faces a dilemma: (1) HAPPO requires independent policies with No Parameter-sharing (NoPS) to guarantee monotonic improvement; (2) Heterogeneous MARL generally requires the Partial Parameter-sharing (ParPS) mechanism to achieve optimal cooperative performance.
Our experiment further reveals that directly combining ParPS with HAPPO causes the \textit{policy updating baseline drift} problem, thereby disrupting HAPPO's convergence.

To address these challenges, we propose the Optimal Marginal Deterministic Policy Gradient (OMDPG) method for heterogeneous MARL, featuring three innovations:
\begin{itemize}
    \item \textbf{Optimal Marginal Contribution Quantification (OMQ)}: Integrating Shapley value and marginal optimization theory, we design the Optimal Marginal Q-value (OMQ) using Q-functions to quantify individual actions' marginal contributions to joint advantages. This replaces sequentially updated $Q_{\psi}^s(s,a_{1:i})$ with parallelizable optimal joint action sequences, resolving the conflict between monotonic improvement and partial parameter-sharing while preserving the MAAD guarantee.

    \item \textbf{Generalized Q Critic (GQC)}: Replacing value functions with Q-functions as critics, we employ pessimistic uncertainty-constrained loss to optimize Q-value estimations across agent types, providing both required OMQ computation inputs and stable actor updating baselines.

    \item \textbf{Centralized Critic Grouped Actor (CCGA) Architecture}: A centralized shared critic ensures accurate global Q-function computation, while grouped parameter-sharing in policy networks captures both heterogeneity and homogeneity.
\end{itemize}


\section{Related Work}
Heterogeneous MARL is very common in real-world scenarios, such as heterogeneous LLM agents for financial sentiment analysis \cite{xing2024designing} and multi-agent robotic collaboration systems \cite{robotic+yoon2019learning,robotics+yang2021can}.
It has emerged as a critical paradigm for solving cooperative tasks involving agents with distinct capabilities and action spaces. Unlike homogeneous systems where agents share identical policies and reward structures, heterogeneous MARL addresses scenarios where agents exhibit diverse behaviors and policies.
Traditionally, the heterogeneity in MARL can be categorized into two classes: \textit{physical} and \textit{behavioral} \cite{Def-HetGPPO-bettini2023heterogeneous,Def-wilson2022evolution,Def-wilson2022performance}.
In GHQ \cite{GHQ-yu2024ghq}, the authors give a formal definition of physical heterogeneity with the usage of \textit{local transition function}. In GAPO \cite{GAPO-yu2025solving}, the authors solve the action semantic conflict in heterogeneous MARL problems.

In general, MARL approaches face unique challenges in maintaining policy monotonicity due to the inherent non-stationarity of multi-agent environments. Policy monotonic improvement ensures that agents' performance increases consistently during training, which is crucial for stable convergence in cooperative settings.
QMIX \cite{rashid2020+qmix-jmlr} introduces monotonic value function factorization through a mixing network with non-negative weights.
MAVEN \cite{mahajan2019maven} enhances QMIX's exploration through latent space perturbations, solving the monotonicity-constrained exploration dilemma.
MEHAML \cite{MEHARL-liu2023maximum} incorporates policy entropy regularization, which guarantees monotonic improvement through mirror descent updates with regret bounds.
OATRPO \cite{OATRPO-chen2024off} extends TRPO to MARL by introducing an off-agent experience sharing mechanism with theoretical guarantees for monotonic improvement.
MAT \cite{MAT-wen2022multi} reformulates MARL as sequence modeling, achieving monotonic improvement through advantage decomposition with theoretical guarantees for linear time complexity in agent count.

\section{Definition and Analysis}

\subsection{Preliminaries}
We adopt the Decentralized Partially Observable Markov Decision Process (Dec-POMDP) \cite{DEC-POMDP+oliehoek2016concise} framework to study heterogeneous cooperative MARL problems. The problem is formalized as a tuple $G=\left\langle S, A, O; \mathbb{P}, \Omega, R; \gamma, N, K, T \right\rangle$, where $N= \{ 1, ..., n_{i}, ..., n  \} $ denotes the finite set of $n$ agents, $K= \{ 1, ..., k_{i}, ...,  k  \} $ represents the finite set of $k$ agent groups and $\gamma \in [0,1)$ is the reward discount factor.

The true environmental state $s \in S$ contains complete system information but remains inaccessible to decentralized actors. At timestep $t \le T$, agent $i \in N$ in group $k_i \in K$ receives local observation $o_i^t$ and selects action $a_i^t \in A_i$ from its available action set $A_i$. All agent actions form the joint action $a^t = (a_1^t, ..., a_n^t) \in A = (A_1, ..., A_n)$, with $a_{1:i}^t = (a_1^t, ..., a_i^t) \in A$ denoting the first $i$ agents' joint action. The environment updates its state through transition function $\mathbb{P} (s^{t+1}|s^{t}, a^t)$ and generates rewards $r^t$ via global reward function $R(s, a^t)$. All local observations $o_1^t, ..., o_n^t$ are generated by observation function $\Omega (s^t, i)$, forming joint observations $\boldsymbol{o}^t = (o_1^t, ..., o_n^t) \in O$.

The observation-action history $\tau^t_i = \cup_1^t \{ (o^{t-1}_i, a^{t-1}_i)\}$ ($t \ge 1$; $\tau^0 = o^0$) represents agent $i$'s local transition tuples before timestep $t$, while $\tau_i$ specifically denotes agent $i$'s complete trajectory over $t \le T$. The experience replay buffer $\mathcal{D} = \cup (\tau, s, r)$ stores batch-sampled data. Network parameters are uniformly denoted as $\theta$ and $\psi$.

\subsection{Problem Analysis}

To achieve monotonic improvement, the MAAD theorem in HAPPO \cite{happo+kuba2021trust} requires sequential updates. To circumvent direct optimization of the sequential advantage function $Adv^{\boldsymbol{\pi}}_{i}(s,a_{1:i-1},a_{i})$, equations (9), (10), and (11) in the HAPPO paper is conducted to approximate $Adv^{\boldsymbol{\pi}}_{i}(s,a_{1:i-1},a_{i})$. This is achieved by sequentially calculating the ratio between old and updated policies for preceding agents:
\begin{equation}
    F_i=\frac{\overline{\pi}_{1:i-1}(a_{1:i-1}|s)}{\pi_{1:i-1}(a_{1:i-1}|s)} \ ,
\end{equation}
which is then multiplied by the global advantage function $Adv_{\boldsymbol{\pi}}(s,a)$ to obtain the approximated sequential advantage for the optimization objective for policy $\pi_i$:
\begin{equation}
\label{ch5-equa-serial-adv}
    Adv^{\boldsymbol{\pi}}_{i}(s,a_{1:i-1},a_{i}) \simeq  F_i \times  Adv_{\boldsymbol{\pi}}(s,a) \ .
\end{equation}

\begin{figure}[htb]
    \centering
	\includegraphics[width=0.99\textwidth]{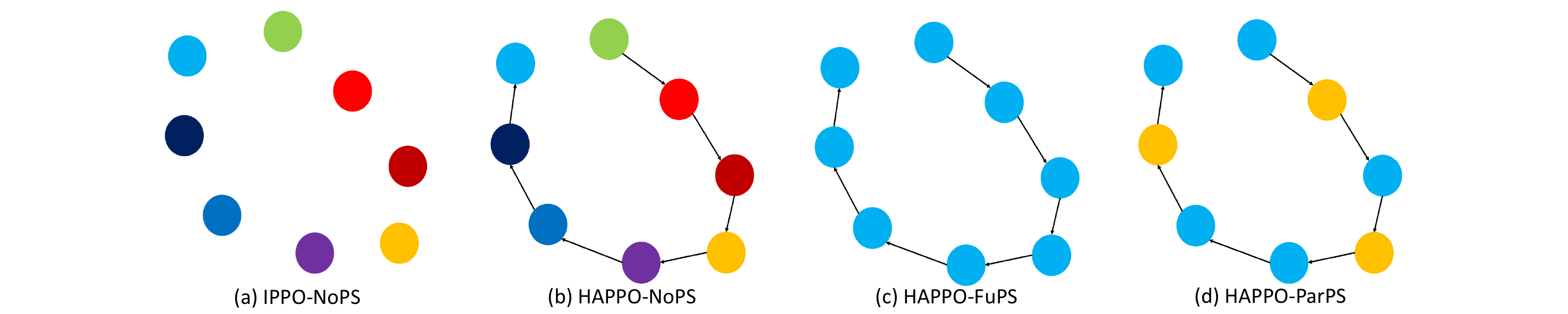}
	\caption{Schematic diagram of parameter sharing and training paradigms in different baseline algorithms.}
	\label{ch5-fig-SMAC-ParPS}
\end{figure}

Applying this method with ParPS induces the \textit{policy updating baseline drift} problem. As illustrated in Fig. \ref{ch5-fig-SMAC-ParPS}, the colored circles represent agents in different groups and black arrows denote MAAD-required sequential updates. In Fig. \ref{ch5-fig-SMAC-ParPS}(d), after updating the first blue agent's parameters, subsequent blue agents' parameters are inadvertently altered via ParPS before their scheduled $F_i$ calculations. This leads to a corrupted $F_3$ calculation when updating the third blue agent after the second yellow agent,
manifesting the policy updating baseline drift phenomenon:
\begin{equation}
    F_3=\frac{\overline{\pi'}_{1:2}(a'_{1:2}|s)}{\pi'_{1:2}(a'_{1:2}|s)} \ne \frac{\overline{\pi}_{1:2}(a_{1:2}|s)}{\pi_{1:2}(a_{1:2}|s)} \ .
\end{equation}

\subsection{Validation Experiments in SMAC Environments}

\begin{figure}[htb]
    \centering
	\includegraphics[width=0.99\textwidth]{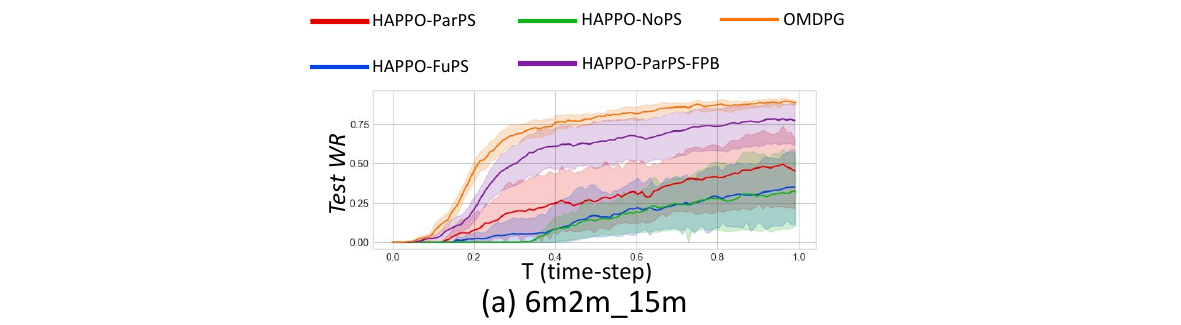}
	\caption{Validation experiments using HAPPO as baseline in SMAC environments.}
	\label{ch5-fig-SMAC-HAPPO}
\end{figure}

As is shown in Fig. \ref{ch5-fig-SMAC-HAPPO}, we evaluate various parameter-sharing paradigms' impacts using HAPPO-NoPS as a baseline. The results are illustrated in Fig. \ref{ch5-fig-SMAC-ParPS}. The green curve represents HAPPO-NoPS, which is the original implementation from \cite{happo+kuba2021trust} that fully preserves MAAD-compliant sequential optimization.

The red and blue curves correspond to HAPPO-ParPS and HAPPO-FuPS (full parameter-sharing), respectively. While benefiting from strategy similarity through parameter-sharing to discover cooperative tactics faster, both fail to converge to high win rates (WR) with large variance oscillations. This confirms that the policy updating baseline drift problem undermines HAPPO's monotonic improvement guarantees.
The purple curve HAPPO-ParPS-FPB introduces a fixed policy baseline (FPB) method, precomputing all "pre-update policies" $\overline{\pi}_{1:i}(a_{1:i}|s)$ before optimization. Although this ensures accurate $\overline{\pi}$ calculation, it neglects sequential update effects and inter-agent policy interactions, resulting in biased $F_i$ estimations. While showing notable WR improvements, it still underperforms the OMDPG algorithm's optimal level.

Our analysis demonstrates the ineffectiveness of the policy updating baseline drift caused by the combination of ParPS and HAPPO. HAPPO-ParPS exhibits local sub-optimality, and the FPB method cannot completely solve the problem.

\section{Methods}
\subsection{Optimal Marginal Q-value}

To adapt HAPPO's sequential update paradigm to value-based methods, Liu et al. \cite{GSE-liu2024solving} proposed using greedy marginal contributions (Shapley Value) as optimization targets for sequential state-action value functions $Q^s$:
\begin{equation}
\label{ch5-equa-GSE-phi}
    \phi_i^*(\tau_i,a_{1:i-1},a_i) = Q_i^c(\tau_i,a_{1:i-1},a_i,a_{i+1:n}^*) - V_i^c(\tau_i,a_{1:i-1},a_{i+1:n}^*) \ ,
\end{equation}
\begin{equation}
\label{ch5-equa-GSE-loss}
    \mathcal{L}=\mathbb{E}_{\pi}\left[\left(Q^s(\tau_i,a_{1:i-1},a_i) - \phi_i^*(\tau_i,a_{1:i-1},a_i)\right)^2\right] \ ,
\end{equation}
where $Q_i^c$ denotes the critic Q-function for agent $i$. By substituting $V_i^c(\tau_i,a_{1:i-1},a_{i+1:n}^*)$ with $Q_i^c(\tau_i,a_{1:i-1},\mathbf{0},a_{i+1:n}^*)$ and replacing the optimal joint action $a_{i+1:n}^*$ with value-greedy joint action $a_{i+1:n}^g$ to achieve monotonic improvement in value decomposition methods.
We first observe that under MAAD theory, the sequential Q function $Q^s_i(s,a_{1:i-1},a_i)$ and sequential advantage function $Adv^s_i(s,a_{1:i-1},a_i)$ are equivalent:
\begin{equation}
\begin{aligned}
    Adv^{s}_{i}(s,a_{1:i-1},a_{i})
    &= Adv^s_{i}(s,a_{1:i}) - Adv^s_{i-1}(s,a_{1:i-1}) \\
    &= \left(Q^s_{i}(s,a_{1:i}) - V(s)\right) - \left(Q^s_{i-1}(s,a_{1:i-1}) - V(s)\right) \\
    &= Q^s_{i}(s,a_{1:i}) - Q^s_{i-1}(s,a_{1:i-1}) \\
    &= Q^s_i(s,a_{1:i-1},a_{i})\ ,
\end{aligned}
\end{equation}
enabling direct optimization using $Q^s_i(s,a_{1:i-1},a_i)$ instead of Equation \ref{ch5-equa-serial-adv}'s approximation. By substituting local observation trajectories $\tau_i$ with global states $s$ in Equations \ref{ch5-equa-GSE-phi} and \ref{ch5-equa-GSE-loss}, we derive the Optimal Marginal Q-value (OMQ):
\begin{equation}
\begin{aligned}
    \phi_i^*(s,a_{1:i})
    &= Q^s_i(s,a_{1:i-1},a_i) \\
    &= Q_i^c(s,a_{1:i-1},a_i,a_{i+1:n}^*) - V_i^c(s,a_{1:i-1},a_{i+1:n}^*) \\
    &= Q_i^c(s,a_{1:i-1},a_i,a_{i+1:n}^*) - Q_i^c(s,a_{1:i-1},\mathbf{0},a_{i+1:n}^*) \ ,
\end{aligned}
\end{equation}
which fundamentally resolves the policy updating baseline drift problem by replacing sequential policy ratios $F_i$ with optimally computed joint actions $(a_{1:i-1},a_i,a_{i+1:n}^*)$.

\subsection{Generalized Q Critic}

To compute and optimize $\phi_i^*(s,a_{1:i})$, we propose the Generalized Q Critic (GQC) framework, which employs a unified Q-function to estimate three key values: $Q_i^c(s,a_{1:n})$, $Q_i^c(s,a_{1:i-1},a_i,a_{i+1:n}^*)$, and $Q_i^c(s,a_{1:i-1},\mathbf{0},a_{i+1:n}^*)$.

Here, $Q_i^c(s,a_{1:n})$ represents the true state-action value from actual sampled trajectories, used for critic network loss computation. The other two values involve out-of-distribution joint actions $(a_{1:i-1},a_{i+1:n}^*)$ not present in real samples, creating estimation challenges analogous to offline RL's out-of-distribution (OOD) problem.

Integrating methods from PBRL \cite{PBRL-bai2022pessimistic} and MATD3 \cite{MATD3-zhang2020td3}, we introduce a Pessimistic Uncertainty Loss (PU) for OOD actions:
\begin{equation}
\begin{aligned}
    \mathcal{U}(s, a_{1:i},a_{i+1:n}^*) &= Std(Q^{C}(s, a_{1:i},a_{i+1:n}^*)) \\
    &= \sqrt{\frac{1}{C_k} \sum_{C_i=1}^{C_k} \left(Q^{C_i}(s, a_{1:i},a_{i+1:n}^*) - \overline{Q^{C}}(s, a_{1:i},a_{i+1:n}^*)\right)^2 } \ ,
\end{aligned}
\end{equation}
where $C_k$ denotes the number of parallel critic networks, and $\overline{Q^{C}}$ is the mean Q-value across all critics. Each critic $C_i \in C_k$ independently samples trajectories $\tau_{C_i} \sim \mathcal{D}$ and maintains separate target networks $Q^{tgt}_{C_i}$.
For real trajectory updates, we have:
\begin{equation}
\label{ch5-equa-true-target}
    y_{C_i}^{true} = r + \gamma \max_a Q^{tgt}_{\psi_{C_i}}\left(s', \tilde{a}' + \epsilon\right) \ ,
\end{equation}
\begin{equation}
\label{ch5-equa-true-loss}
    \mathcal{L}(\psi_{C_i}^{true}) = \mathbb{E}_{(\tau,s) \sim \mathcal{D}} \left[ \left( Q_{\psi_{C_i}}(s, a) - y_{C_i}^{true} \right)^2 \right] \ ,
\end{equation}
where $\tilde{a}^{t+1}$ is generated by policy network $\pi_{\theta_i}$ with truncated Gaussian noise $\epsilon \sim \mathcal{N}(0, \sigma)$.
For OOD updates, we replace optimal actions $a_{i+1:n}^*$ with policy-greedy joint actions and yield the PU-constrained targets:
\begin{equation}
\label{ch5-equa-policy-greedy-action}
    a_{i+1:n}^g = \left\{ \max_{a_{i+1}} \pi_{i+1}^{\theta_{i+1}}(\tau_{i+1}), \ldots, \max_{a_n} \pi_n^{\theta_n}(\tau_n) \right\} \ ,
\end{equation}
\begin{equation}
\label{ch5-equa-pseudo-target}
    y_{C_i}^{PU} = Q_{\psi_{C_i}}^{tgt}(s, a_{1:i},a_{i+1:n}^g) - \beta \mathcal{U}(s, a_{1:i},a_{i+1:n}^g) \ ,
\end{equation}
\begin{equation}
\label{ch5-equa-pseudo-loss}
    \mathcal{L}(\psi_{C_i}^{PU}) = \mathbb{E}_{(\tau,s) \sim \mathcal{D}} \left[ \left( Q_{\psi_{C_i}}(s, a_{1:i},a_{i+1:n}^g) - y_{C_i}^{PU} \right)^2 \right] \ ,
\end{equation}
where $\beta$ controls PU constraint strength. Target networks update via soft updates to prevent Q-value overestimation.
The complete GQC loss combines both components, where $\lambda_{PU}$ balances loss contributions:
\begin{equation}
    \mathcal{L}(\psi_{C_i}^{GQC}) = \mathcal{L}(\psi_{C_i}^{true}) + \lambda_{PU} \mathcal{L}(\psi_{C_i}^{PU}) \ .
\label{ch5-equa-total-GQC-loss}
\end{equation}

\subsection{Algorithm Implementation}

The Optimal Marginal Deterministic Policy Gradient (OMDPG) framework employs deterministic policy gradient methods to construct actor networks, implementing ParPS by partitioning actors into $K$ groups: $(\pi_{\theta_1},...,\pi_{\theta_k})$. Each actor group receives group-specific observation trajectories $\tau_{G_i}$ to generate corresponding group actions $a_{G_i}$, forming the Centralized Critic Grouped Actor (CCGA) architecture. This enables effective balance between intra-group policy similarity and inter-group cooperation.

\begin{figure}[htb]
    \centering
	\includegraphics[width=0.99\textwidth]{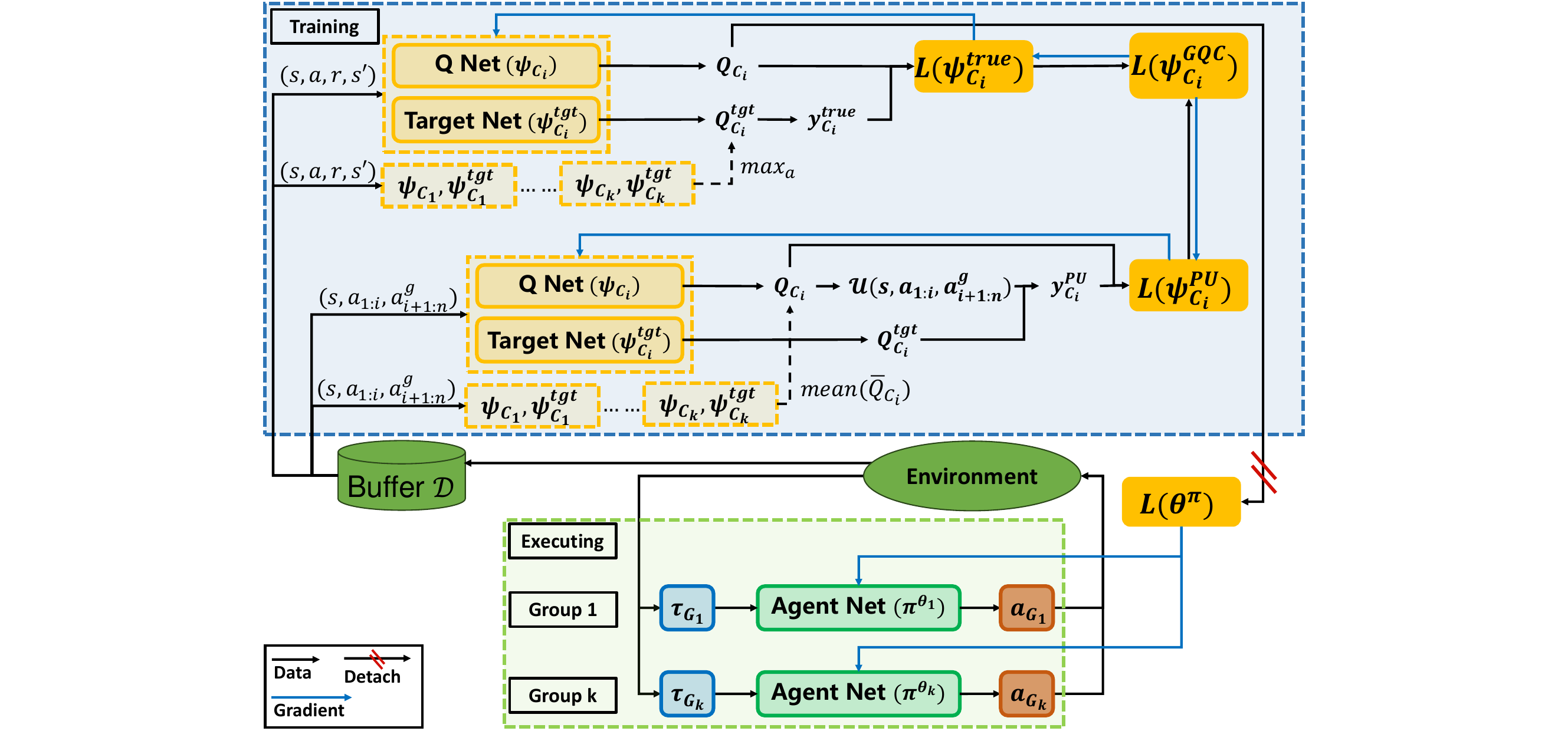}
	\caption{Architecture of the OMDPG algorithm.}
	\label{ch5-fig-OMDPG-Arch}
\end{figure}

The temporal difference (TD) loss function for actor networks aims to maximize the action values estimated by critic networks. Based on pessimistic Q-value estimation, the loss function is formulated as:
\begin{equation}
\begin{aligned}
\mathcal{L}(\pi^{TD}_{\theta_i}) &= -\mathbb{E}_{\tau \sim \mathcal{D}, a \sim \pi} \left[ \min_{C_i \in C_k} Q_{\psi_i}^s(s,a_{1:i}) \right] = -\mathbb{E}_{\tau \sim \mathcal{D}, a \sim \pi} \left[ \min_{C_i \in C_k} \phi_{\psi_i}^*(s,a_{1:i}) \right] \\
&= -\mathbb{E}_{\tau \sim \mathcal{D}, a \sim \pi} \left[ \min_{C_i \in C_k} \left( Q_{\psi_i}(s,a_{1:i-1},a_i,a_{i+1:n}^g)-Q_{\psi_i}(s,a_{1:i-1},\mathbf{0},a_{i+1:n}^g) \right)\right],
\label{ch5-equa-policy-loss}
\end{aligned}
\end{equation}
where $a_{i+1:n}^g$ denotes policy-greedy joint actions, and $\min_{C_i \in C_k} Q_{\psi_i}$ indicates using the minimum Q-value among all $C_k$ critics for PU loss calculation and also as the actor optimization target. Fig. \ref{ch5-fig-OMDPG-Arch} illustrates the OMDPG architecture.



\section{Experiments and Analysis}  

\subsection{Comparative Experiments in SMAC Heterogeneous Cooperative Environments}

\begin{figure}[htb]
    \centering
	\includegraphics[width=0.99\textwidth]{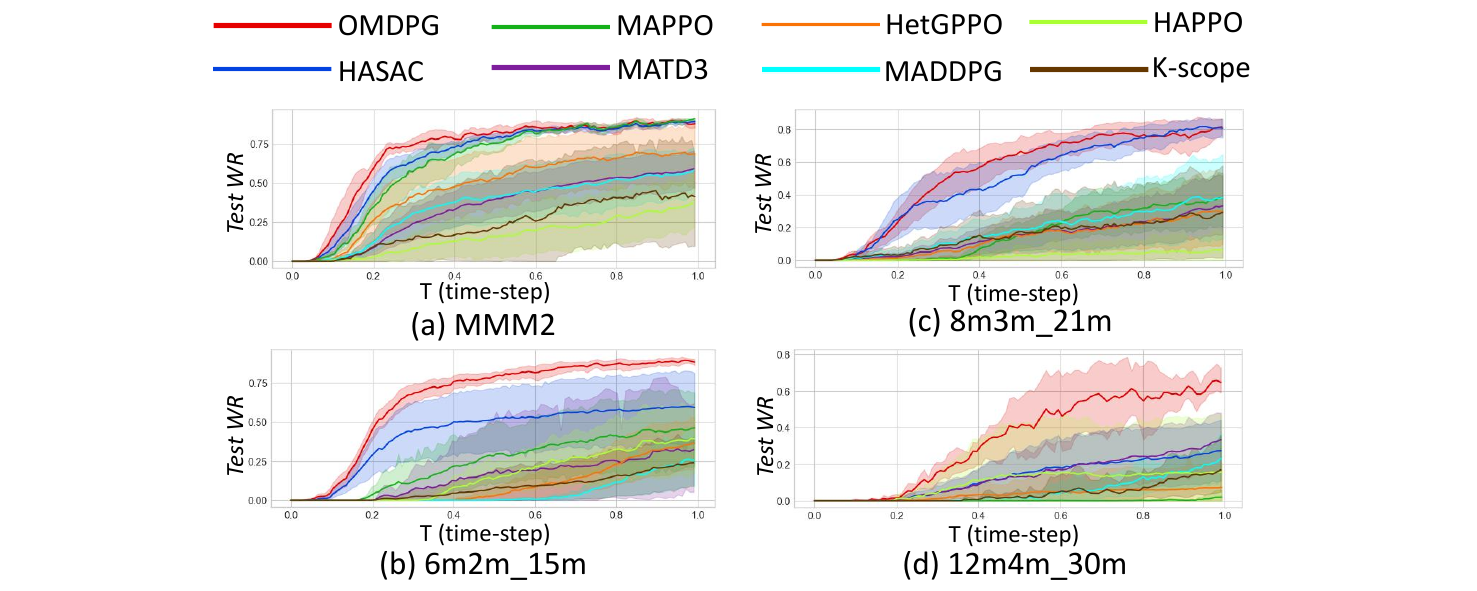}
	\caption{Comparative Experimental Curves in SMAC Environments}
	\label{ch5-fig-All-SMAC}
\end{figure}

Fig. \ref{ch5-fig-All-SMAC} and Table \ref{ch5-table-policy-based-SMAC} summarize the experimental results in SMAC heterogeneous cooperative environments.

(1) \textbf{Simple Scenario: Testing all algorithms on the original asymmetric heterogeneous map MMM2}. In Fig. \ref{ch5-fig-All-SMAC}(a), OMDPG, HASAC \cite{HASAC-2024liumaximum}, and MAPPO \cite{mappo+yu2021surprising} converge to a winning rate (WR) of 0.9 within approximately 6M training steps. HetGPPO \cite{Def-HetGPPO-bettini2023heterogeneous} achieves $WR\approx0.7$ with high variance. MATD3 \cite{MATD3-zhang2020td3} and MADDPG \cite{MPE-lowe2017multi} attain $WR\approx0.6$ with relatively low variance. HAPPO \cite{happo+kuba2021trust} and Kaleidoscope \cite{li2024kaleidoscope} only reach $WR\approx0.4$ with maximal variance.\par

(2) \textbf{Medium Scenario: Increased number of allied units and enemy units for difficulty balancing}. In Fig. \ref{ch5-fig-All-SMAC}(b), OMDPG outperforms all baselines with $WR\approx0.9$ (minimal variance). HASAC achieves $WR\approx0.6$ but exhibits high variance. MAPPO reaches $WR\approx0.4$ with significant variance, while other methods fail below $WR=0.4$. HAPPO and MATD3 show similar learning curves, but MATD3 has greater variance. HetGPPO achieves the smallest variance yet underperforms MATD3, Kaleidoscope, and MADDPG.
In Fig. \ref{ch5-fig-All-SMAC}(c), both OMDPG and HASAC achieve $WR\approx0.8$ with the lowest relative variance. Other methods plateau below $WR=0.4$ with comparable performance and high variance, indicating their failure to learn effective policies in scenario (c). OMDPG demonstrates the fastest convergence speed and optimal monotonic improvement, conclusively validating the effectiveness.
The failure of other baseline methods indicates their disadvantage in cooperating among heterogeneous agents.\par

\begin{table}[htb]
\centering
\caption{Algorithm Performance Comparison Across SMAC Scenarios}
\label{ch5-table-policy-based-SMAC}
\begin{tabular}{*{5}{c}}
    \toprule
    Algorithm & MMM2 & 6m2m\_15m & 8m3m\_21m & 12m4m\_30m \\
    \midrule
    OMDPG & $0.89_{(0.10)}$ & $\textbf{0.87}_{(0.07)}$ & $\textbf{0.82}_{(0.09)}$ & $\textbf{0.65}_{(0.08)}$ \\
    HASAC & $0.90_{(0.06)}$ & $0.58_{(0.42)}$ & $0.80_{(0.13)}$ & $0.29_{(0.34)}$ \\
    MAPPO & $\textbf{0.93}_{(0.03)}$ & $0.47_{(0.38)}$ & $0.38_{(0.24)}$ & $0.02_{(0.05)}$ \\
    MATD3 & $0.65_{(0.34)}$ & $0.38_{(0.25)}$ & $0.35_{(0.34)}$ & $0.37_{(0.28)}$ \\
    HetGPPO & $0.64_{(0.43)}$ & $0.37_{(0.25)}$ & $0.32_{(0.32)}$ & $0.08_{(0.23)}$ \\
    MADDPG & $0.65_{(0.43)}$ & $0.26_{(0.20)}$ & $0.41_{(0.37)}$ & $0.30_{(0.24)}$ \\
    HAPPO & $0.45_{(0.32)}$ & $0.39_{(0.28)}$ & $0.11_{(0.13)}$ & $0.15_{(0.30)}$ \\
    Kaleidoscope & $0.23_{(0.39)}$ & $0.31_{(0.24)}$ & $0.33_{(0.35)}$ & $0.20_{(0.23)}$ \\
    \bottomrule
\end{tabular}
\end{table}


(3) \textbf{Complex Scenario: Significant unit quantity escalation}. In Fig. \ref{ch5-fig-All-SMAC}(d), only OMDPG achieves $WR\approx0.6$, significantly surpassing all baselines. Among baselines, only MATD3 and HASAC achieve $WR\approx0.3$ with high variance. Other methods fail below $WR=0.2$, indicating complete policy learning failure.
The failure of baseline methods demonstrates that out-of-distribution joint optimal action groups $(a_{1:i-1},a_{i+1:n}^g)$ become critical in complex scenarios for policy optimization. The usage of GQC methods with PU loss is necessary to mitigate their impact for accurate policy optimization and a monotonic improvement guarantee. \par

In summary, the OMDPG method successfully implements ParPS and sequential policy updates in SMAC environments, achieving superior monotonic improvement and asymptotic performance across all difficulty levels compared to baseline methods.
Other baseline methods lack a proper cooperating mechanism for heterogeneous agents while also lack the monotonic improvement guarantee, which finally leads to their poor performance in SMAC experiments.

\subsection{Comparative Experiments in MAMuJoCo Environments}

\begin{figure}[htb]
    \centering
	\includegraphics[width=0.99\textwidth]{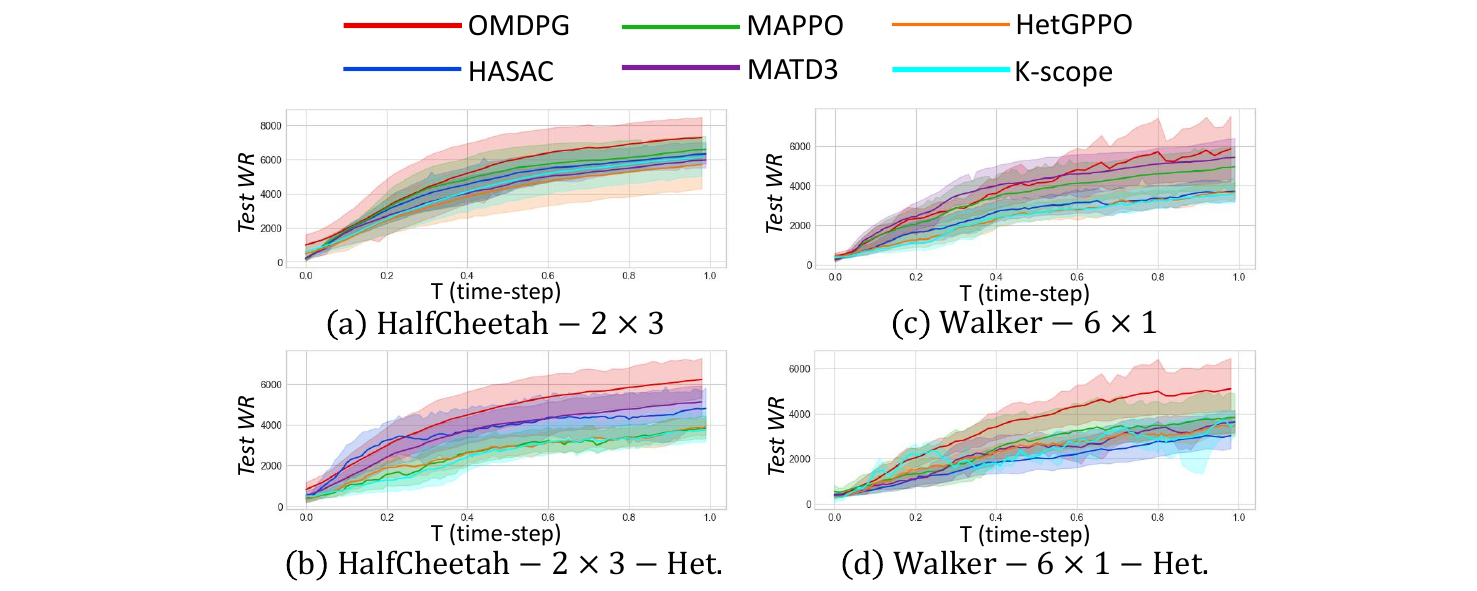}
	\caption{Comparative Learning Curves in MAMuJoCo Environments}
	\label{ch5-fig-All-MAMUJOCO}
\end{figure}

To construct more heterogeneous MAMuJoCo scenarios, we introduce two new environments by modifying joint gear ratios (gear): HalfCheetah-Het. and Walker-Het. (where "Het." denotes Heterogeneous).
For ParPS, HalfCheetah's hind/fore legs and Walker's left/right legs are divided into two groups. Fig. \ref{ch5-fig-All-MAMUJOCO} and Table \ref{ch5-table-policy-based-mamujoco} present the comparative experimental results.

From the figures and tables, OMDPG achieves superior performance in all 4 scenarios. Comparing subfigures (a)-(b) and (c)-(d) in Fig. \ref{ch5-fig-All-MAMUJOCO}, OMDPG demonstrates the smallest performance degradation despite increased environmental complexity from gear modifications. The success of OMDPG reaffirms the necessity of GQC methods with PU loss for mitigating out-of-distribution action impacts in complex environments. Consistent with SMAC findings, OMDPG achieves optimal monotonic improvement and coordination across all scenarios. \par


\begin{table}[htb]
\centering
\caption{Algorithm Performance Comparison in MAMuJoCo Environments}
\label{ch5-table-policy-based-mamujoco}
\begin{tabular}{*{5}{c}}
    \toprule
    Algorithm & HalfCheetah & HalfCheetah-Het. & Walker & Walker-Het. \\
    \midrule
    OMDPG & $\textbf{7923.5}_{(769.8)}$ & $\textbf{7161.2}_{(957.5)}$ & $\textbf{6414.0}_{(2201.3)}$ & $\textbf{5719.4}_{(1623.1)}$ \\
    HASAC & $6340.6_{(651.5)}$ & $4902.0_{(1078.5)}$ & $3752.0_{(893.5)}$ & $4196.0_{(1441.0)}$ \\
    MAPPO & $6604.8_{(535.4)}$ & $3872.1_{(716.4)}$ & $5032.3_{(1082.8)}$ & $3827.6_{(905.3)}$ \\
    MATD3 & $6018.8_{(389.7)}$ & $6156.1_{(1384.1)}$ & $5490.6_{(850.4)}$ & $3686.9_{(519.6)}$ \\
    HetGPPO & $6809.7_{(1104.5)}$ & $4070.6_{(803.5)}$ & $3775.3_{(683.4)}$ & $3620.1_{(799.1)}$ \\
    Kaleidoscope & $7132.1_{(1065.3)}$ & $3759.5_{(821.1)}$ & $3629.8_{(525.8)}$ & $4044.7_{(224.6)}$ \\
\bottomrule
\end{tabular}
\end{table}

\subsection{Ablation Studies}
\subsubsection{1. Ablation study about the OMQ and GQC module.}

\begin{figure}[htb]
    \centering
	\includegraphics[width=0.99\textwidth]{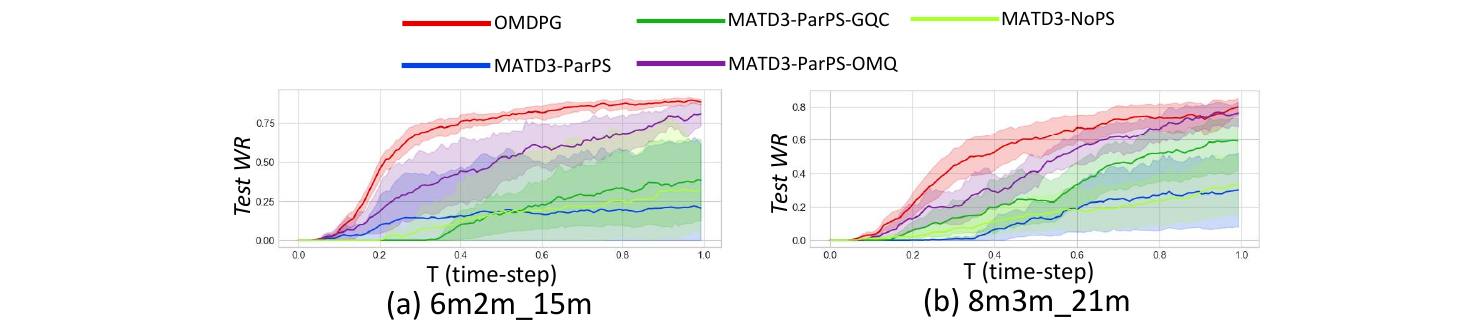}
	\caption{Ablation Experiments in SMAC Environments}
	\label{ch5-fig-SMAC-Ablation}
\end{figure}

To validate the individual contributions of OMQ and GQC methods in addressing the policy updating baseline drift problem arising from combining ParPS with sequential optimization, we conduct ablation experiments in (a) 6m2m\_15m and (b) 8m3m\_21m scenarios. Using MATD3-NoPS as the baseline, we establish three ablation groups.
The MATD3-ParPS-GQC group uses 5 GQC functions, while MATD3-ParPS-OMQ uses 2 GQC functions. Results are shown in Fig. \ref{ch5-fig-SMAC-Ablation}. \par

Overall, the OMQ method demonstrates greater performance enhancement contributions, whereas standalone GQC usage yields unstable outcomes. Both (a) and (b) show comparable WR between MATD3-ParPS and MATD3-NoPS baseline, indicating that ParPS alone fails to deliver significant WR improvements.
The MATD3-ParPS-OMQ group exhibits superior WR curves with lower variance compared to other groups, confirming OMQ's effectiveness.
MATD3-ParPS-GQC is equivalent to using out-of-distribution joint actions to enhance Q-network exploration, but increases variance and network complexity. This explains the slow WR improvement and maximal variance observed in the MATD3-ParPS-GQC group in (a). In the more complex (b) scenario with larger agent populations, MATD3-ParPS-GQC shows greater performance gains, indicating heightened potential in challenging environments.

\subsubsection{2. Ablation study about the $C_k$, $\lambda_{PU}$, and the computational cost.}
In order to determine the optimal values of hyper-parameters $C_k$ and $\lambda_{PU}$, we conduct several groups of experiments. The results of $C_k$ experiments are summarized in Table \ref{table-CK}, and the results of $\lambda_{PU}$ experiments are summarized in Table \ref{table-lambda-PU}. The best values of hyper-parameters are in bold font.
The OMDPG runner and the SMAC environment are implemented with multi-threaded and parallel methods. Therefore, we quantitatively analyze the computational cost of the change of $C_k$ in OMDPG and the HAPPO and MATD3 methods using the actual computing time of the algorithms in MMM2. The results in Table \ref{table-Computational Cost} prove that the computational time costs of OMDPG are less than HAPPO and MATD3. Even though increasing $C_k$ causes more computational cost, considering the improvement of WR, the cost is generally affordable. All of the experiments are conducted on the same workstation with the AMD Ryzen 7950x CPU, Nvidia RTX 4090 GPU, and Kingston 128~G RAM with 20 sampling threads and 5 learning threads. \par

\begin{table}
\centering
\caption{Results of Ablation Study on the Hyper-parameter $C_k$.}
\label{table-CK}
\centering
\begin{tabular}{cccccc}
\toprule
Map Name & $C_k=2$ & $C_k=3$ & $C_k=4$ & $C_k=\textbf{5}$ & $C_k=6$ \\
\midrule
MMM2 & $0.75_{(0.10)}$ & $0.79_{(0.09)}$ & $0.84_{(0.11)}$ & $\textbf{0.89}_{(0.10)}$ & $0.88_{(0.14)}$ \\
6m2m\_15m & $0.79_{(0.09)}$ & $0.81_{(0.08)}$ & $0.82_{(0.12)}$ & $\textbf{0.87}_{(0.07)}$ & $0.87_{(0.11)}$ \\
8m3m\_21m & $0.75_{(0.11)}$ & $0.80_{(0.09)}$ & $0.82_{(0.10)}$ & $\textbf{0.82}_{(0.09)}$ & $0.80_{(0.10)}$ \\
12m4m\_30m & $0.60_{(0.12)}$ & $0.61_{(0.11)}$ & $0.62_{(0.11)}$ & $\textbf{0.65}_{(0.08)}$ & $0.64_{(0.10)}$ \\
\bottomrule
\end{tabular}
\end{table}

\begin{table}
\centering
\caption{Results of Ablation Study on the Hyper-parameter $\lambda_{PU}$.}
\label{table-lambda-PU}
\centering
\begin{tabular}{cccccc}
\toprule
Map Name & $\lambda_{PU}=0.01$ & $\lambda_{PU}=0.03$ & $\lambda_{PU}=0.05$ & $\lambda_{PU}=0.07$ & $\lambda_{PU}=\textbf{0.1}$ \\
\midrule
MMM2 & $0.74_{(0.08)}$ & $0.77_{(0.10)}$ & $0.76_{(0.11)}$ & $0.82_{(0.12)}$ & $\textbf{0.89}_{(0.10)}$ \\
6m2m\_15m & $0.72_{(0.09)}$ & $0.76_{(0.12)}$ & $0.83_{(0.12)}$ & $0.86_{(0.10)}$ & $\textbf{0.87}_{(0.07)}$ \\
8m3m\_21m & $0.71_{(0.16)}$ & $0.79_{(0.18)}$ & $0.81_{(0.14)}$ & $0.83_{(0.12)}$ & $\textbf{0.82}_{(0.09)}$ \\
12m4m\_30m & $0.61_{(0.20)}$ & $0.64_{(0.14)}$ & $0.62_{(0.16)}$ & $0.63_{(0.11)}$ & $\textbf{0.65}_{(0.08)}$ \\
\bottomrule
\end{tabular}
\end{table}

\begin{table}
\centering
\caption{Analysis of Computational Cost of OMDPG and other algorithms in MMM2.}
\label{table-Computational Cost}
\centering
\begin{tabular}{ccccc}
\toprule
Items & OMDPG($C_k=2$) & OMDPG($C_k=5$) & HAPPO & MATD3 \\
\midrule
Computing Time (h) & 18.9 & 19.6 & 14.2 & 10.5 \\
Converged Time-step (M) & 7.2 & 8.5 & / & / \\
Converged Time (h) & 12.6 & 14.8 & / & / \\
\bottomrule
\end{tabular}
\end{table}

\section{Conclusion}  

In this paper, we focus on improving the monotonic improvement under the combination of grouped parameter sharing with sequential policy updates in heterogeneous MARL. We first reveal the policy updating baseline drift problem, which negatively impacts policy learning. To address this issue, we propose the Optimal Marginal Deterministic Policy Gradient (OMDPG) algorithm, featuring two core components: the Optimal Marginal Q (OMQ) function and Generalized Q Critic (GQC).
OMQ replaces the sequential policy update ratio $F_i$ with the optimal marginal contribution value $\phi_{\psi}^*(s,a_{1:i})$ under joint optimal actions, achieving significant performance gains. To mitigate the impact of out-of-distribution joint optimal actions, we introduce GQC with pessimistic uncertainty (PU) loss to quantify and control induced uncertainties.
Finally, following the Centralized Critic Grouped Actor (CCGA) architecture, we develop the OMDPG algorithm.
Comparative and ablation experiments in SMAC and MAMuJoCo heterogeneous cooperative environments validate OMDPG's efficacy.

In the future, we intend to explore heterogeneous MARL problems in large-scale scenarios and environments. The scalability problem is crucial for the MARL algorithm application in real-world scenarios, especially with the help of the fast-developing LLM techniques. Future improvements and developments could focus on integrating communication or LLM techniques for better adaptability of heterogeneous MARL methods in complex scenarios.


%
%
%
\makeatletter
\renewcommand\@biblabel[1]{#1.}
\makeatother
\bibliography{main}
%




\end{document}